\documentclass{IEEEtran}

\usepackage{cite}
\usepackage{algorithmic}
\usepackage{amsmath, amssymb, amsfonts}
\usepackage{graphicx}
\usepackage{textcomp}
\usepackage[english]{babel}
\usepackage{soul} 
\usepackage[utf8]{inputenc}
\usepackage{float}
\usepackage{hyperref}

\makeatletter
\def\ps@IEEEtitlepagestyle{%
  \def\@oddfoot{
      \scriptsize \parbox{\textwidth}{\hfil This work has been submitted to the IEEE for possible publication. Copyright may be transferred without notice, after which this version may no longer be accessible. \hfil}%
  }
}
\makeatother

\title{DeepLight: A Sobolev-trained Image-to-Image Surrogate Model for Light Transport in Tissue}
\author{Philipp Haim, Vasilis Ntziachristos, Torsten En{\ss}lin, Dominik J\"ustel
    \thanks{This project has received funding from the European Research Council (ERC) under the European Union’s Horizon Europe research and innovation programme under grant agreement No 101041936 (EchoLux) and from the Bavarian Ministry of Economic Affairs, Energy and Technology (StMWi) (DIE-2106-0005// DIE0161/02, DeepOpus) (Corresponding Author: Dominik J\"ustel, e-mail: dominik.juestel@tum.de).}
    \thanks{Philipp Haim, Dominik Jüstel, and Vasilis Ntziachristos are with the Chair of Biological Imaging, Central Institute for Translational Cancer Research (TranslaTUM), School of Medicine and Health \& School of Computation, Information and Technology, Technical University of Munich, 81675 Munich, Germany,
        and also with the Institute of Biological and Medical Imaging, Bioengineering Center, Helmholtz Zentrum München, 85764 Neuherberg, Germany.
    }
    \thanks{Philipp Haim and Dominik J\"ustel are also with the Institute of Computational Biology, Helmholtz Zentrum Munich, 85765 Neuherberg, Germany,
    and also with the Institute of AI for Health, Helmholtz Zentrum Munich, 85764 Neuherberg, Germany.}
    \thanks{Philipp Haim is also with the Ludwig-Maximilians-Universität München, 80539 Munich, Germany}
    \thanks{Torsten En{\ss}lin is with the Max-Planck-Institute for Astrophysics, 85741 Garching, Germany,
        also with the German Centre for Astrophysics, 02826 G\"orlitz, Germany}
}

\begin{document}
\maketitle
\begin{abstract}
    In optoacoustic imaging, recovering the absorption coefficients of tissue by inverting the light transport remains a challenging problem. %
    Improvements in solving this problem can greatly benefit the clinical value of optoacoustic imaging. %
    Existing variational inversion methods require an accurate and differentiable model of this light transport. %
    As neural surrogate models allow fast and differentiable simulations of complex physical processes, they are considered promising candidates to be used in solving such inverse problems. %
    However, there are in general no guarantees that the derivatives of these surrogate models accurately match those of the underlying physical operator. %
    As accurate derivatives are central to solving inverse problems, errors in the model derivative can considerably hinder high fidelity reconstructions. %
    To overcome this limitation, we present a surrogate model for light transport in tissue that uses Sobolev training to improve the accuracy of the model derivatives. %
    Additionally, the form of Sobolev training we used is suitable for high-dimensional models in general. %
    Our results demonstrate that Sobolev training for a light transport surrogate model not only improves derivative accuracy but also reduces generalization error for in-distribution and out-of-distribution samples. %
    These improvements promise to considerably enhance the utility of the surrogate model in downstream tasks, especially in solving inverse problems. %
\end{abstract}


\section{Introduction}
\begin{figure*}[!ht]
    \begin{center}
        \includegraphics[width=\textwidth]{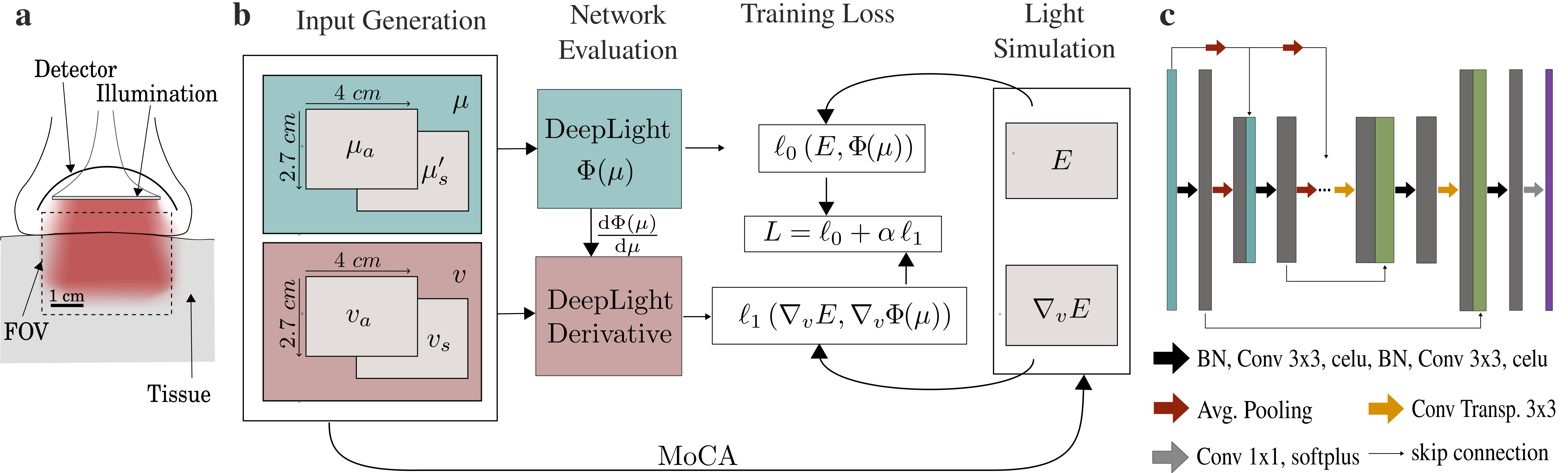}
    \end{center}
    \caption{
        (a) Illustration of the optoacoustic measurement setup. %
        A probe head with an illumination source is placed on the patient’s skin. %
        Short pulses of laser light excite acoustic pressure waves, which can be recorded by acoustic detectors. %
        (b) Schematic depiction of the data generation and training procedure. %
         Synthetically generated absorption and scattering coefficients ($\mu_a$, $\mu_s^\prime$), and randomly sampled derivative directions ($v_a$, $v_s$) are used as input for the Monte Carlo ray tracer MoCA to compute the absorbed energy ($E$) as well as the corresponding directional derivative ($\nabla_v E$). %
         The training loss ($L$) of the DeepLight network includes a term that penalizes errors in $E$ ($\ell_0$) and a term for the derivative accuracy ($\ell_1$) scaled by the hyper-parameter $\alpha$. %
        (c) Sketch of the network architecture. %
        An adapted UNet architecture was implemented. %
        Notably, after each downsampling step, a downsampled copy of the input was concatenated to the inputs of the next layer. %
        BN: Batch Norm, Conv: Convolution, Avg.: Average, Conv Transp.: Transpose convolution
    }
    \label{fig:schematic}
\end{figure*}

Optoacoustic (OptA) tomography is a medical imaging modality that allows non-invasive imaging of optical contrast deep within tissue, achieving resolution comparable to acoustic imaging methods. %
The method operates on the principle of the OptA effect, where illuminating an absorbing molecule (i.e. chromophore) with a short light pulse results in the excitation of a pressure wave, which can be detected by ultrasound transducers~\cite{Ntziachristos2010,Lutzweiler2013}. %
However, a key challenge persists in accurately recovering the spatial distribution of absorbers, with the inversion of the light transport in tissue being the main difficulty~\cite{Knieling2025}.

Variational optimization methods address this problem by iteratively fitting an initial guess to the data using gradient-based optimizers. %
Achieving high-quality reconstructions with variational methods necessitates an accurate and differentiable model of the measurement process. %
For practicality, the model must be sufficiently fast, as it may need to be evaluated thousands of times per reconstruction. %
In this study, our focus is constructing a model of light transport in biological tissue that fulfills these constraints.\par%

The radiative transfer equation (RTE) is widely regarded as the gold standard for modeling light transport in tissue. %
Solving the RTE is in general computationally expensive, making it ill-suited for the use in inverse problems. %
Since our focus is on the absorbed energy field within a highly turbid medium, we can make use of a commonly adopted approximation of the RTE, the diffusion equation. %
The diffusion equation can be solved significantly faster than the RTE and is valid in highly scattering media, such as the biological tissue typically imaged with OptA modalities~\cite{Cheong1990}. %
Nevertheless, this simplification presents certain challenges. %
Accurate approximation requires an effective diffusive source, which differs from the actual illumination and is not known in general~\cite{Tarvainen2005}. %
Furthermore, the approximation breaks down in the coupling medium and the epidermis, where ballistic light transport dominates~\cite{Hielscher1998,Aydin2002}. %
As those superficial layers can exert non-negligible effects on the subsequent light transport, incorrect modeling those layers can negatively impact reconstruction results in all deeper layers.\par

A more accurate computation of absorbed energy can be achieved with Monte-Carlo (MC) methods~\cite{Cox2012,Banerjee2010,Fang2009}. %
However, MC approaches are computationally expensive and derivatives challenging to evaluate. %
Both of those problems can be addressed by employing a neural surrogate model, i.e., a neural network trained to reproduce accurate MC simulations of light transport. %
Such a network provides an accurate, fast and differentiable model for an OptA imaging modality~\cite{Mishra2021}. %
However, conventional supervised learning does not guarantee that the surrogate model derivatives accurately match those of the underlying operator that is learned~\cite{Cocola2020}. %
Since variational optimization relies on those model derivatives, mismatches can negatively impact the accuracy and fidelity of the reconstruction. %
A surrogate model with accurate derivatives is needed for use in inverse problems and could improve the recovery of accurate chromophore density distributions in OptA imaging.\par

An established method for learning functions and their derivatives with a neural network is Sobolev training. %
This approach defines the training loss on a Sobolev space, penalizing both network predictions and network derivatives up to a specified order~\cite{Czarnecki2017}. %
In this study, we will only focus on first order derivatives. %
For models with vector-valued inputs or outputs, Sobolev training then requires the computation of the error of the model Jacobian. %
Although the full computation of the Jacobian becomes infeasible for high dimensional problems, it can be estimated stochastically. %
Networks trained with such a loss function have been shown to accurately learn both the function values and its derivatives, and often exhibit a lower generalization loss. %
When used as surrogate models for reconstructions, Sobolev-trained networks outperformed networks trained solely on function values~\cite{Laurent2019,Tsay2021}. %
 However, Sobolev training has not yet been demonstrated on a high-dimensional network, such as an image-to-image network.\par

Previously, OptA imaging surrogate models for light transport were presented~\cite{Rix2023}. %
Those models are intended to allow fast computation of synthetic data, rather than for use in reconstruction. %
Consequently, the authors did not consider the derivatives of their surrogate model.\par

To summarize, there is currently neither a light transport surrogate model that utilizes the potential benefits of Sobolev training, nor work that demonstrates the feasibility of Sobolev training for high dimensional models. %
We hypothesize that including a single sampling direction of the Jacobian for each training image is sufficient to result in improvements in the network performance. %
In this work, we demonstrate that this approach to Sobolev training for high dimensional models improves the derivative accuracy and generalization ability in an image-to-image light transport surrogate model.

\section{Methods}
We trained a neural network to reproduce simulated images of absorbed energy distributions using absorption and reduced scattering coefficients as inputs. %
The loss function included a term that penalized inaccuracies in the directional derivatives of the network. %
The simulation setup was modelled on an existing OptA imaging system, with Figure~\ref{fig:schematic}a illustrating the measurement setup.

\subsection{Architecture}
A convolutional neural network architecture based on UNet~\cite{UNet} was employed. %
Figure~\ref{fig:schematic}c shows a schematic representation of the model. %
The network receives as input 2D absorption coefficients ($\mu_a$) and reduced scattering coefficients ($\mu_s^\prime$), where $\mu_s^\prime=\frac{\mu_s}{1-g}$ with $\mu_s$ representing the scattering coefficient and $g$ representing scattering anisotropy. %
The final layer of the network outputs the fluence $F$, from which the absorbed energy  can be subsequently computed as
\begin{equation}
    E = F\,\mu_a.
    \label{eq:fluence}
\end{equation}
Since our quantity of interest is $E$, our full model $\Phi(\mu$) outputs the product of $F$ and $\mu_a$.
\par
Before the input was passed to the network, a third channel with the depth of each pixel is added to the inputs. %
Although this third channel remains constant within the model and contains redundant information, its inclusion was justified by the strong influence of depth on fluence. %
Adding this third channel allowed the network to account for this relationship more easily and additionally helped to break the translation symmetry present in the convolutions at the encoder blocks.
\par
Each encoder block of the network performs two sets of batch normalizations and convolutions, followed by a down sampling step via average-pooling. %
The decoder blocks upsamples the input, concatenates it with the output from the corresponding encoder block and convolves the result. %
Unlike the original UNet architecture, we added a skip connection combined with average-pooling from the input channels to each encoder block. %
Each encoder block could use this (in principle redundant) information to learn to represent light transport at the length scales from the downscaled inputs. %
The output of the last decoder block is then subjected to a 1x1 convolution that returns a single channel. %
Multiplying this channel with the input $\mu_a$ yields the final output of the network.

\subsection{Training}
We used a loss function defined on the Sobolev space of square integrable functions whose first weak derivative exists and is itself square integrable.
In our scenario, we want to evaluate the difference between the ground truth operator that computes the absorbed energy $E$ and the output of the surrogate model $\Phi$.
For any sample of optical coefficients $\mu=(\mu_a,\, \mu_s)$, the loss takes the form:
\begin{equation}
    \ell(E, \mu)\ =\ \ell_0(E,\ \Phi(\mu))\ + \alpha\, \ell_1(D_\mu E,\ D_\mu \Phi(\mu)),
    \label{eq:sobolev_image_loss}
\end{equation}
where $\ell_0$ and $\ell_1$ are norms, commonly the $L^2$ or Frobenius norm, $\alpha$ is a scalar weighting parameter and $D_\mu$ is the first weak derivative with respect to $\mu$. %
\par
For vector-valued inputs and outputs, as is the case here, the derivative becomes a Jacobian matrix. %
The dimensionality of this Jacobian is determined by multiplying the dimensions of the arguments by the dimensions of the output. %
Computing this Jacobian for an image-to-image network is computationally infeasible. %
However, it is possible to estimate the norm of the Jacobian by computing the expectation value of the Jacobian applied to randomly sampled vectors, an approach referred to as stochastic Sobolev-training~\cite{Czarnecki2017}. %
We can then write the derivative loss term $\ell_1$ as the expectation value over random vectors $v$ of a loss $\ell_1^*$ on directional derivatives. %
The loss then reads:
\begin{align}
    \ell\left(E, \mu\right)\ =\ \ell_0(E,\ \Phi(\mu))\ + \alpha\left\langle \ell_1^*\big(\nabla_v E,\ \nabla_v \Phi(\mu)\big)\right\rangle_{v \sim V}
    \label{eq:stoch_sobolev_loss}
\end{align}
where $\nabla_v$ denotes the directional derivative for $v$, comprised of vectors of change in absorption ($v_a$) and scattering ($v_s$), and $V$ is the probability distribution of the random vectors $v$. %
\par
The expectation value can be approximated by evaluating $\ell_1^*$ for a finite number of vectors. %
As the evaluation of directional derivatives is computationally expensive, there is a tradeoff between generating a larger, more varied dataset and obtaining a better approximation of the Jacobian loss. %
We used an extreme case of this approximation by using only a single vector sample. %
Due to this extremely sparse sampling of the Jacobian, we chose the distribution $V$ to be biased towards vectors we expect to be relevant for reconstructions. %
Its implementation is detailed in section 3. %
\par
The total loss between the ground truth operator and the surrogate model is then defined as the sum of $\ell$ over pairs of $\mu$ and $v$:
\begin{align}
    L\ =\ \sum_{(\mu, v)}\Big(\ell_0(E,\ \Phi(\mu))\ + 
        \alpha\ell_1^*\big(\nabla_v E,\ \nabla_v \Phi(\mu)\big)\Big)
    \label{eq:stoch_sobolev_loss_detailed}
\end{align}
\par
\begin{figure}
    \begin{center}
        \includegraphics[width=0.48\textwidth]{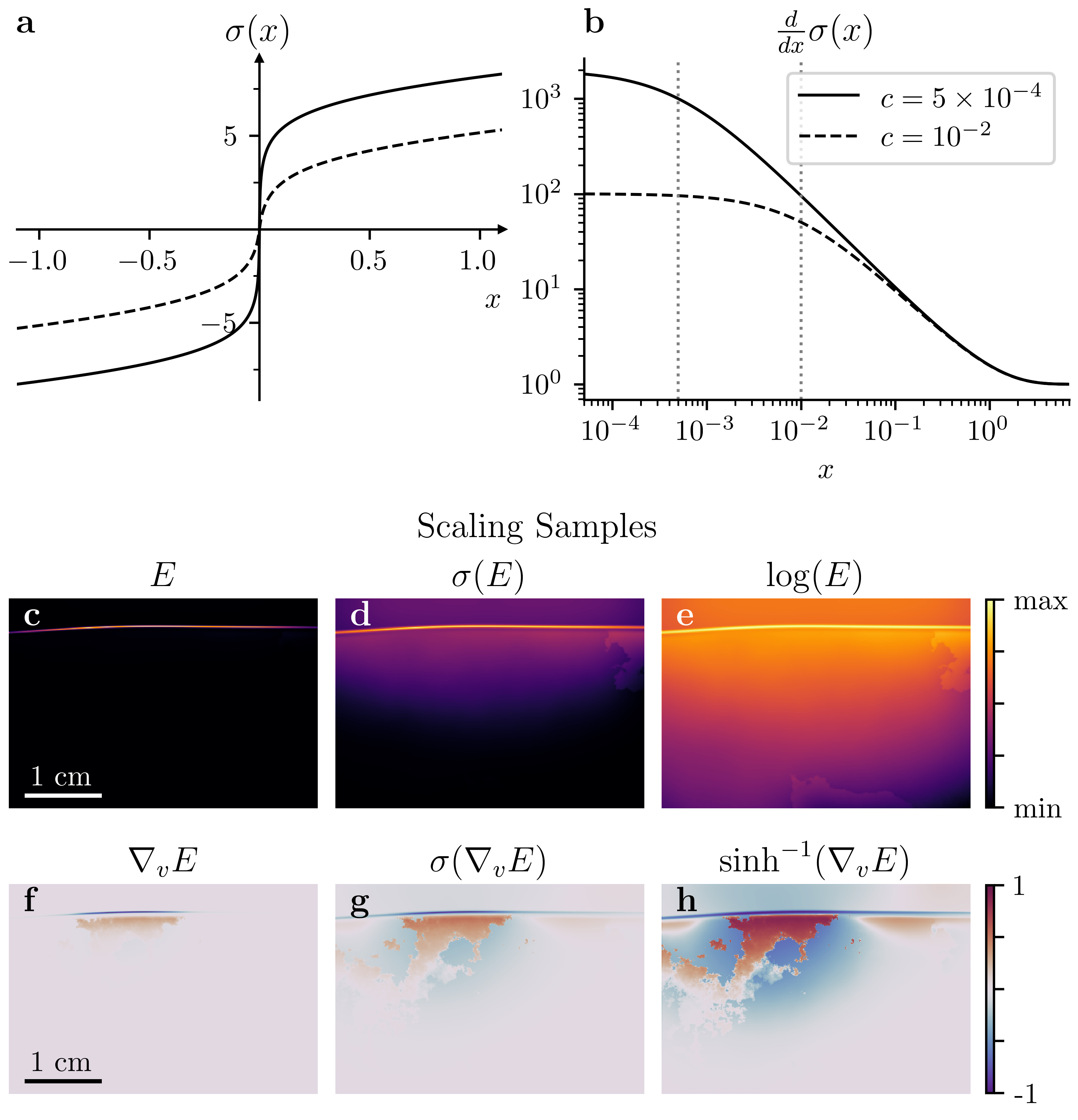}
    \end{center}
    \caption{
        (a) Nonlinear scaling function $\sigma$, (b) its derivative, and (c-h) their effects on absorbed energy ($E$) and directional derivative ($\nabla_v E$) in synthetic tissue samples compared to linear and logarithmic scaling. %
        The function is plotted for $a=1$ and two cutoff-scale values $c$, which are also marked as vertical lines in panel (b). %
        The derivatives approach a value of 1 for input values larger than $a$, corresponding to linear scaling. %
        Between $c$ and $a$, logarithmic scaling is observed. %
        For inputs with magnitude below $c$, the derivative flattens out again, approaching a constant derivative when inputs are close to 0. %
        Absorbed energy with (c) linear, (d) $\sigma$ and (e) logarithmic scaling. %
        Directional derivatives with (f) linear, (g) $\sigma$ and (h) symmetric-logarithmic scaling. %
    }
    \label{fig:nl}
\end{figure}
To choose appropriate loss functions $\ell_0$ and $\ell_1$, the high dynamic range that needs to be accurately learned by the network must be considered. %
A common approach to handle a high dynamic range of strictly positive values is to train the network on the logarithm of those values, optimizing for a small relative error in the whole image. %
As features such as the skin or superficial vessels can absorb orders of magnitude more energy than lower absorbing or deeper structures, the network error in the most absorbing regions could become larger than the absorbed energy in fainter regions of the image. %
Such a discrepancy could be detrimental for reconstructions with variational methods, potentially leading to a complete failure to reconstruct fainter regions. %
\par
To address those issues, we introduce the scaling function $\sigma$ defined as follows:
\begin{equation}
    \sigma_{a, c}(x) = \text{sign}(x)\log\left(\frac{e^{(\vert x \vert + c) / a} - 1}{e^{c/a} - 1}\right)
    \label{eq:nl}
\end{equation}
The scaling behavior of this function can be divided into three distinct regimes. %
For values larger than $a$, $\sigma_{a, c}$ exhibits asymptotic linear scaling. %
Within the range ($c$, $a$), the scaling is logarithmic, whereas for values smaller than $c$, the function exhibits linear scaling. %
For readability we will omit the parameters $a$ and $c$ when referring to $\sigma$, unless required for clarity. %
Figure~\ref{fig:nl}a shows the function itself, and Figure~\ref{fig:nl}b shows the derivative of $\sigma$, demonstrating its scaling behavior. %
By adjusting parameters $a$ and $c$, it is possible to select values for which the training loss is sensitive to the absolute prediction error ($x>a$ and $x<c$) or the relative error ($c<x<a$). %
Since $\sigma_{a, c}$ and its derivative are well defined for negative numbers, it can also be applied to directional derivatives. %
Figure~\ref{fig:nl}c shows the absorbed energy for a tissue sample with moderately high skin absorption. %
In this example, absorption by the skin line is predominant, completely overshadowing features located deeper within the tissue. %
In contrast, logarithmic scaling (as seen in Figure~\ref{fig:nl}e) reveals structures at all depths of the tissue. %
However, the deepest features are not relevant in this scenario, as they are beyond the detection capabilities of current OptA measurement devices. %
Furthermore, intensity variations in the skin line are overshadowed by the much larger variations between shallow and deep tissue regions. %
The image scaled with $\sigma$, shown in Figure~\ref{fig:nl}d, presents a compromise between linear and logarithmic scaling. %
Although it provides contrast for deeper tissues, this improved contrast is limited to the range where signal detection is feasible.
Intensity variations along the skin line also remain evident. %
The directional derivatives, shown in Figure~\ref{fig:nl}f-h, exhibit similar behavior. %
Since the derivatives can stretch to negative numbers, we replaced the logarithm with $\sinh\left(\frac{\cdot}{c}\right)$, which behaves logarithmically for input values with magnitude larger than $c$ and linearly otherwise. %
Again, linear scaling provides only superficial contrast, while $\sinh^{-1}$ scaling leads to saturation effects in areas with high sensitivity with respect to changes in absorption or scattering. %
The use of $\sigma$ mitigates those issues. %
\par
Using this transformation, each loss term in \eqref{eq:stoch_sobolev_loss_detailed} can be expressed as follows:
\begin{equation}
	\ell(f(x), g(x)) = \left\Vert (\sigma_{a, c} \circ f)(x) - (\sigma_{a^\prime,c^\prime} \circ g)(x) \right\Vert_2
\end{equation}
The parameters $a$ and $c$ were chosen independently for the operator and derivative loss to account for the different scales and depth scaling behaviors of those two quantities. %
Thus, we refer to their respective scaling functions for $E$ and $\nabla_v E$ with their respective parameters as $\sigma^E(\cdot)$ and $\sigma^\nabla(\cdot)$. %
The complete loss is then expressed as follows:
\begin{equation}
    \ell = \left\Vert \sigma^E(E) - \sigma^E(\Phi(\mu))\right\Vert_2 + \alpha \left\Vert \sigma^\nabla(\nabla_v E) - \sigma^\nabla(\nabla_v\Phi(\mu))\right\Vert_2
    \label{eq:loss}
\end{equation}

\subsection{Data Generation}
\begin{figure}
    \begin{center}
        \includegraphics[width=0.49\textwidth]{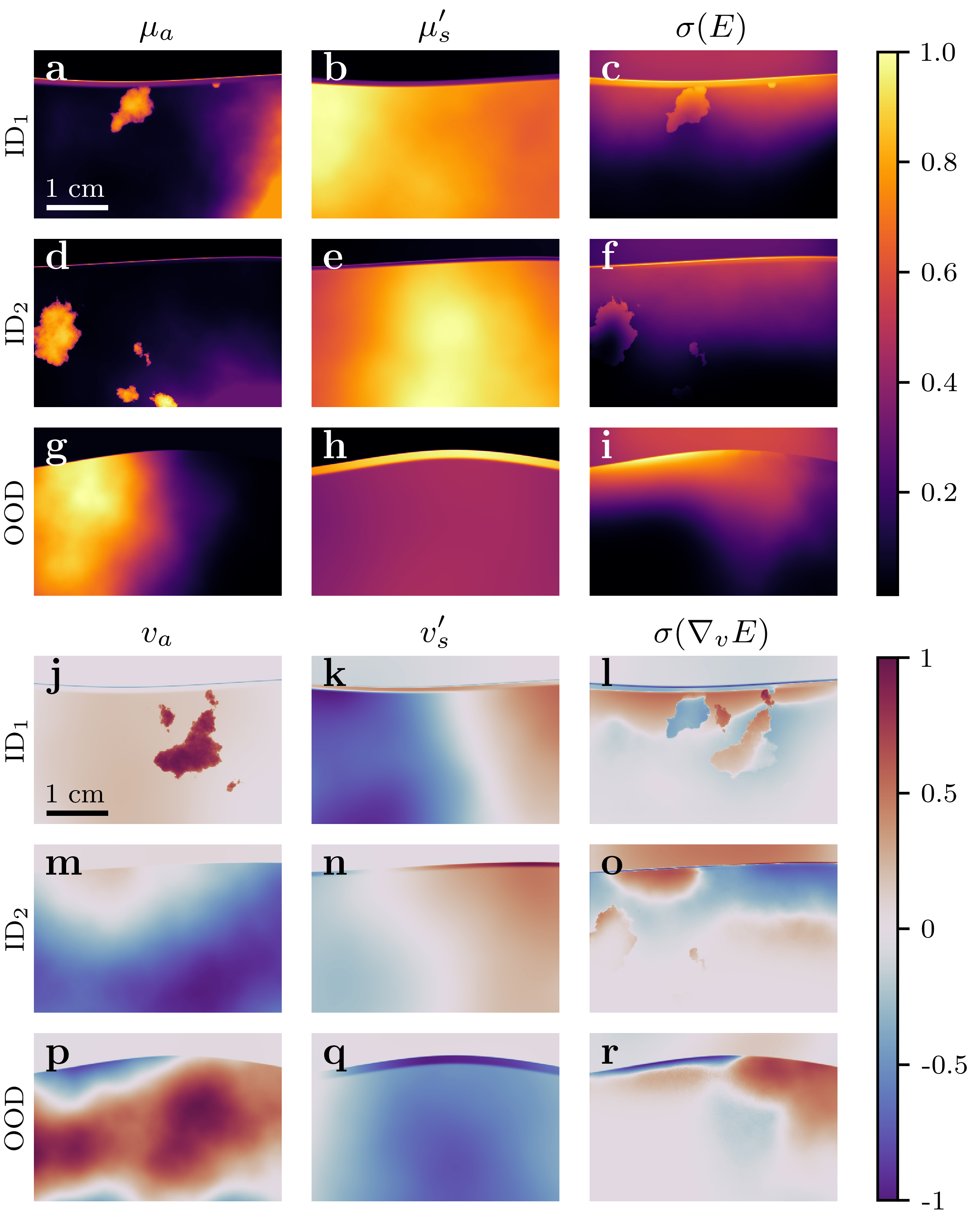}
    \end{center}
    \caption{
        Normalized dataset samples. %
        Panels (a-f) show representative samples of absorption and reduced scattering coefficients ($\mu_a$, $\mu_s^\prime$), as well as the corresponding absorbed energies ($E$) generated by the two in-distribution (ID$_1$ and ID$_2$) generators. %
        $E$ was scaled by $\sigma$ as defined in~\eqref{eq:nl}. %
        Panels (g-i) show $\mu_a$, $\mu_s^\prime$ and $\sigma(E)$ for the out-of-distribution (OOD) generator. %
        In Panels (j-l), samples of the change vectors in absorption ($v_a$), scattering ($v_s^\prime$) and corresponding scaled directional derivative ($\sigma(\nabla_v E)$) from the ID$_1$ generator are shown. %
        Panels (m-o) and (p-r) similarly depict $v_a$, $v_s^\prime$ and $\sigma(\nabla_v E)$ for the ID$_2$ generator and the OOD generator.
    }
    \label{fig:samples}
\end{figure}
We constructed a data generator for absorption and scattering coefficients that replicates the value ranges and structural characteristics found in human tissue. %
Within these constraints, the generator was designed with considerable flexibility to prevent the network from overfitting on specific structures in the training data. %
Specifically, we implemented a tissue mimicking generator that constructs each image from four components:
\begin{enumerate}
    \item Coupling medium with low $\mu_a$ and $\mu_s$ and small spatial variations in those coefficients.
    \item A skin line following a smooth, randomly generated curve with a thin layer mimicking the epidermis. %
    The absorption of this layer spans skin tones on the Fitzpatrick scale from I to V~\cite{Else2023}. %
    Approximately half of the samples included an additional dermis layer with lower values for $\mu_a$ and $\mu_s$. %
   \item A tissue region under the skin with smoothly varying $\mu_a$ and $\mu_s$. %
    The typical magnitude and variabilities of those coefficients were informed by literature values~\cite{Jacques2013}. %
    A minimum absorption was also enforced for this layer based on known tissue properties.
    \item Localized areas with high absorption and low spatial variations to simulate vessels. %
    Their absorption properties are aligned to those of blood.
\end{enumerate}
The spatial variations were generated with a model for a Gaussian processes presented in~\cite{Arras2022}. %
All generated coefficients were constrained to be positive via a component specific nonlinear transformation. %
Anisotropy values for the coupling medium, epidermis, dermis and the remaining tissues were set based on literature values, without variations between samples. %
We will refer to this tissue mimicking generator as the in-distribution (ID) sampler. %
\par
To assess the generalization capabilities of the trained network on out-of-distribution (OOD) samples, we constructed a more generic data generator. %
This sampler uses the same coupling and skin line generator as the ID sampler. %
There were however no epidermis or dermis layers placed at the skin line, nor vessel-like structures in the tissue region. %
Furthermore, we introduced larger variations in $\mu_a$ in the tissue, allowing areas to exhibit almost no absorption. %
We will refer to these samples as generic samples. %
\par
For sampling derivative directions $v_a$ and $v_s$, we used the same generators without a positivity constraint. %
The in-distribution (ID) samples for training and validation consisted of tissue-mimicking samples with tissue-mimicking derivative directions (ID$_1$), as well as tissue-mimicking samples with generic derivative directions (ID$_2$). %
OOD samples consisted of generic samples with generic derivative directions.
\par
Since light transport is fundamentally volumetric, absorbed energy must also be computed as volumetric quantity. %
For the dimension perpendicular to the imaging plane, constant optical properties given by each in-plane pixel were used. %
The volumetric absorbed energy was then mapped back onto two dimensions by summing over the third dimension. %
Figure~\ref{fig:samples} shows representative samples of the generated optical coefficients and corresponding absorbed energy.

\subsection{Monte-Carlo Light Transport}
We needed to compute both absorbed energy densities and their derivatives to generate our training, validation and test datasets. %
Some existing MC frameworks implement adjoint MC or perturbation MC for vector-Jacobian or Jacobian-vector product computation~\cite{Yao2018}. %
However, those differentiation approaches are not well suited for our scenario, which involves both a large number of parameters ($\mu_a$ and $\mu_s$ in each voxel) and numerous detectors ($E$ in each voxel). %
Using such algorithms to generate datasets for Sobolev-training would approximately double the computational cost compared to using datasets containing only absorbed energies. %
To address this limitation, we implemented a custom MC simulator MoCA, capable of simultaneously computing absorbed energy and a directional derivative. %
MoCA’s implementation is based on~\cite{Prahl1989} and~\cite{ElHafi2021}, and its derivative computation follows~\cite{Lataillade2002}. %
MoCA simulates the distribution of absorbed energy within a volume, given the absorption, scattering and anisotropy coefficients. %
Directional derivatives (i.e. Jacobian-vector products) can be computed for changes in both $\mu_a$ and $\mu_s$. %
MoCA is designed to compute derivatives in scenarios involving many parameters and detectors, as is the case in OptA tomography. %
Its applications extend beyond the scope of this work, including sensitivity analyses of OptA signals in response to varying biomarker concentrations.
The code is available at \url{https://github.com/juestellab/MoCA}.\par

\subsection{Validation}
We generated a dataset with 2048 training and 256 validation samples. %
Half of the training and validation samples were generated with the ID$_1$ generator, the other half with the ID$_2$ generator. %
Each sample consists of a discretized $\mu_a$ and $\mu_s^\prime$ as network inputs, $E$ computed with MoCA, as well as derivative directions $v_a$, $v_s^\prime$ and their corresponding $\nabla_v E$. %
For the operator loss, the scale parameter $c$ was set to 1, and for the derivative loss, $c=10$ was used. %
Both loss terms used $a=10^4$. %
With this configuration, only the skin line and strong superficial absorbers remained within the linear regime of $\sigma$.
\par
The optimizer used was ADAM with weight decay (ADAMW)~\cite{Loshchilov2017}. %
During the first 10 training epochs, $\alpha$ was increased linearly from 0 to 0.1, and was then kept constant for the subsequent epochs. %
The learning rate was controlled by a cosine scheduler with one warm restart resulting in two training periods lasting 100 and 200 epochs. %
After the network was trained for the full 300 epochs, the state with the lowest validation operator loss was selected as the result. %
The derivative loss was not used to choose the best performing model in order to keep the decision criteria independent of the Sobolev weight $\alpha$.
To analyze the effect of the Sobolev loss, we also trained a baseline model with Sobolev weight $\alpha=0$, excluding derivatives from the training.
\par
As the stochastic nature of ADAM can lead to different performance outcomes depending on the initialization of the network, we performed the training 5 times with differently seeded random-number generators. %
Finally, the model with the lowest validation operator loss was chosen.
\par
We generated a test dataset $T$ using the generators ID$_1$, ID$_2$ and OOD, each with 128 samples. %
To validate network performance, we evaluated the $L^2$ loss of the absorbed energy transformed by $\sigma$ for each sample in the test set. %
These resulting values were rescaled by the mean $L^2$ norm of the rescaled energy, setting the absolute scale of the errors. %
This error was computed for both the baseline model $\Phi^B$ and the Sobolev-trained DeepLight model $\Phi^D$.
\begin{align}
    d_E(\Phi) &= \frac{\left\Vert \sigma^E(E) - \sigma^E(\Phi(\mu_a, \mu_s^\prime)) \right\Vert_2}{\left\langle \Vert \sigma^E(E^*) \Vert_2\right\rangle_{E^* \in T}}\\
    d_\nabla(\Phi) &= \frac{\left\Vert \sigma^\nabla(\nabla_v E) - \sigma^\nabla(\nabla_v \Phi(\mu_a, \mu_s^\prime)) \right\Vert_2}{\left\langle\Vert \sigma^\nabla(\nabla_v E^*)\Vert_2\right\rangle_{\nabla_v E^* \in T}}
    \label{eq:eval_loss}
\end{align}
We further computed the relative gain of the DeepLight network as:
\begin{equation}
    G_{\{E, \nabla\}} = \frac{d_{\{E, \nabla\}}(\Phi^B) - d_{\{E, \nabla\}}(\Phi^D)}{d_{\{E, \nabla\}}(\Phi^B)}.
    \label{eq:rel_gain}
\end{equation}
\par
Lastly, we analyzed the error of the two networks as a function of depth under the skin line. %
We defined the start of the skin line as the first pixel where scattering surpassed 10~cm$^{-1}$. %
We then computed the average error of a pixel at each depth in the test dataset. %
We also computed the relative gain of the depth dependent error following the definition in~\eqref{eq:rel_gain}. %
\par
The code for the data generation, training and validation of DeepLight can be found at \url{https://github.com/juestellab/DeepLight}.

\section{Results}
\begin{figure*}[!ht]
    \begin{center}
        \includegraphics[width=\textwidth]{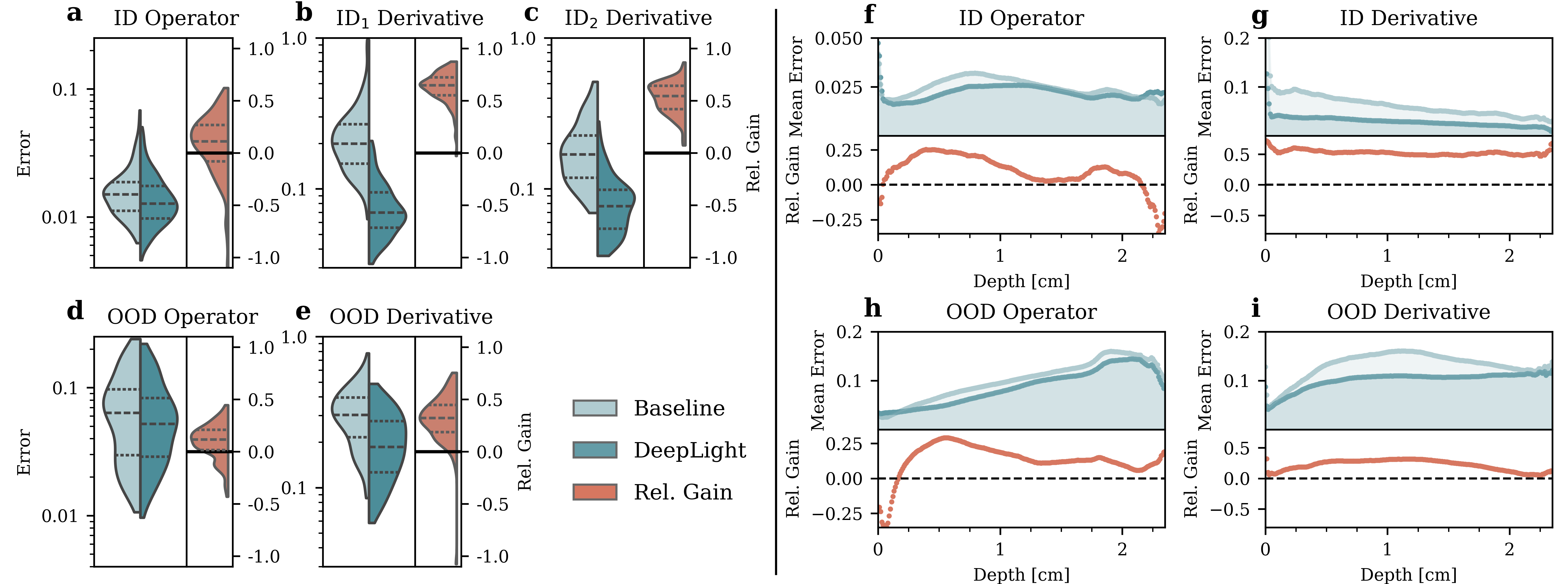}
    \end{center}
    \caption{
        Quantitative comparison of prediction accuracy over the test sets. %
        For each sample, we computed the $L^2$ norm of the prediction error for the scaled absorbed energy and its derivative using the baseline and DeepLight model. %
        To compare errors between test sets, we rescaled the errors by the average $L^2$ norm of the corresponding test set. %
        (a-c) Error distribution for the in-distribution (ID) samples using the ID$_1$ and ID$_2$ generators and the relative gain of the DeepLight model compared to the baseline model. %
        (d) and (e) Error distribution and gain for the out-of-distribution (OOD) samples. %
        (f-i) Average operator and derivative error with respect to pixel depth relative to the skin for both models and the relative gain observed for the ID and OOD samples, respectively. %
        Rel.: relative.
    }\label{fig:results}
\end{figure*}
We found that Sobolev training reduced the network error and improved the derivate accuracy for both ID and OOD samples. %
To assess the effect of Sobolev training, we trained two models with identical architectures, but used different loss functions. %
The baseline model was trained with an $L^2$ loss on the accuracy of the absorbed energy. %
In contrast, our DeepLight model added an additional term that penalized inaccuracies in the directional derivatives of the model. %
The inclusion of this additional loss term results in what is referred to as stochastic Sobolev-training. %
To quantitatively assess the performance of the two networks, the $L^2$ error for each sample, transformed with the scaling function $\sigma$, was computed. %
Our results indicate that the Sobolev-trained Deep-Light model demonstrates superior accuracy compared to the baseline model on both the ID and OOD dataset. %
The DeepLight model also substantially decreased the derivative error. %
Figure~\ref{fig:results} shows the sample-wise error distributions in the test set for the baseline and DeepLight models, as well as their relative gains. %
Figure~\ref{fig:results}a shows that the DeepLight model reduced the error in 68\% of the samples, with a median improvement of 11\%. %
Figure~\ref{fig:results}b and c show substantial decreases in the errors from different derivative samplers used for the ID$_1$ and ID$_2$ generators, showing median relative gains of 62\% and 52\%, respectively. %
Notably, ID$_1$ saw an approximately 80\% decrease in the maximum error. %
Figure~\ref{fig:results}d shows that the error reduction for OOD samples is comparable to the ID samples, with a median relative gain of 12\%, and 78\% of samples having lower errors with the DeepLight model. %
However, the overall error for OOD samples is noticeably higher than those for ID samples, with the median OOD error being approximately 4 times greater than the median ID error. %
Lastly, Figure~\ref{fig:results}e shows that Sobolev training also leads to a reduction in error in the OOD derivatives, with a median relative gain of 32\%.\par %
Analysing the error of the networks with respect to depth relative to the skin, we found that the DeepLight model showed the greatest improvements at depths of around 5~mm. %
The improvement of the derivative accuracy showed less depth dependence, in particular for ID samples. %
For this analysis, we only considered depths of up to 2.3~cm, as stochastic error from the MC simulation starts to dominate the absorbed energy beyond this point, prohibiting meaningful results. %
This threshold encompasses over 99.8\% of all tissue-like pixels in the ID and OOD datasets. %
Figure~\ref{fig:results}f shows that the DeepLight model consistently exhibits a lower error than the baseline model from 0.05 to 2.15~cm below the skin surface, and is only outperformed by the baseline model near the skin line and very deep in tissue. %
Figure~\ref{fig:results}g shows the depth dependent derivative error for the combined ID$_1$ and ID$_2$ samples. %
Over the whole depth, the relative gain is approximately constant at just over 50\%. %
Figure~\ref{fig:results}h presents the depth analysis for the OOD dataset. %
Similar to the ID samples, the baseline model produces a lower error than the DeepLight model in superficial layers up to 2~mm. %
Below this depth, the DeepLight model again demonstrates lower error, outperforming the baseline model. %
Figure~\ref{fig:results}i depicts the OOD derivative error over depth. %
Here we observed the highest gain at a depth between 0.5~cm and 1.5~cm at approximately 25\%. %
Close to the skin line the gain was close to 0, however the absolute error of both models was also much lower than deeper in tissue. %
Overall, we observed that the DeepLight model produced a flatter error curve as a function of depth. %
This implies that the DeepLight model performs more consistently over depth than the baseline model, which appears to sacrifice accuracy in depth in favour of accuracy in the superficial layers.\par
\begin{figure}
    \begin{center}
        \includegraphics[width=0.5\textwidth]{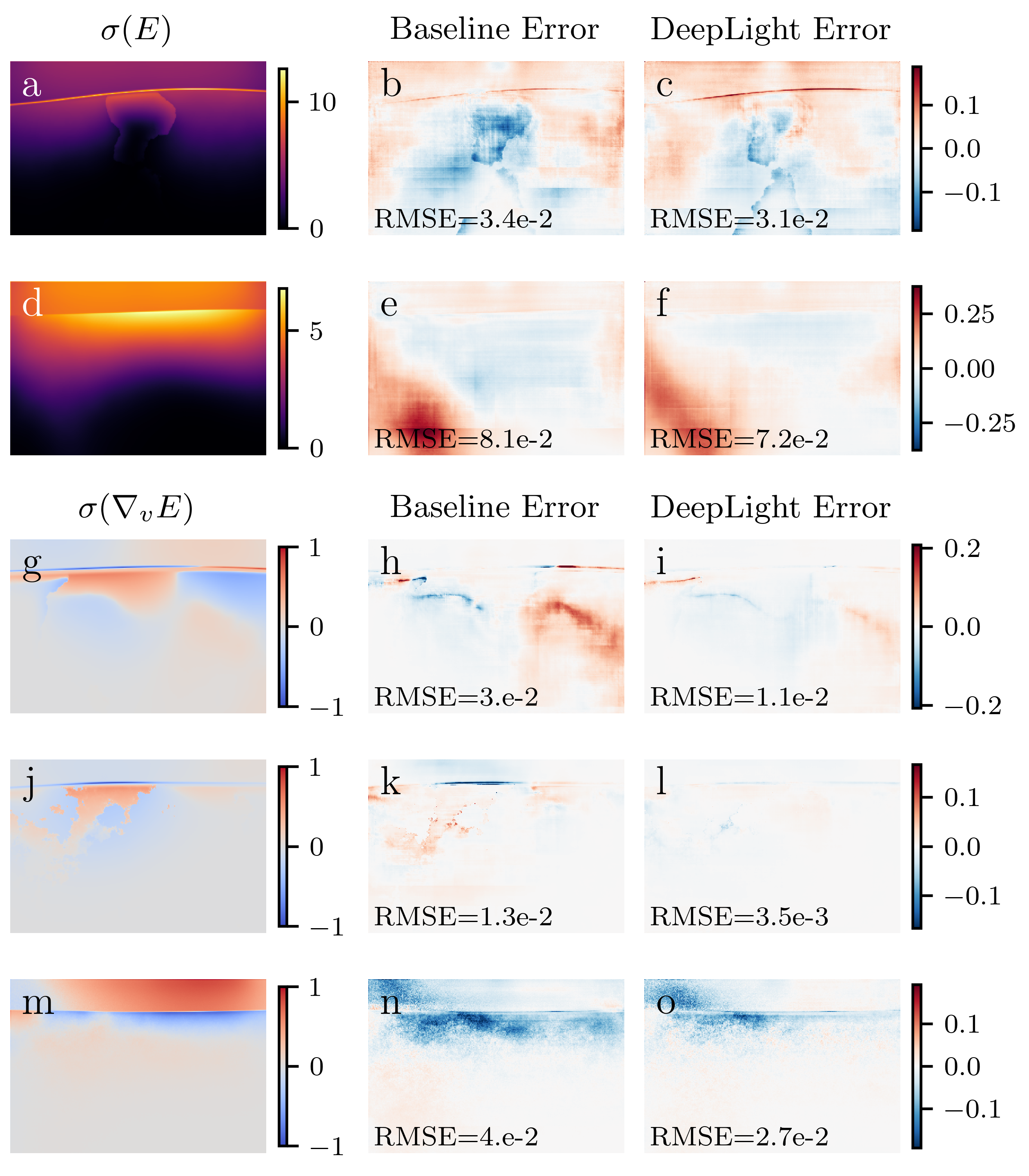}
    \end{center}
    \caption{
    Qualitative samples of predicted absorbed energy $E$ and directional derivatives $\nabla_v E$ with baseline prediction error and DeepLight prediction error. %
    The sample with the median relative gain of the DeepLight model $\Phi_D$ compared to the baseline model $\Phi_B$ is shown. %
    The color range of the difference plots is constrained to the 99.9$^\mathrm{th}$ percentile of the pixel wise errors. %
    (a-c) shows a representative in-distribution sample, (d-f) an out-of-distribution sample. %
    (g-i) and (j-l) show samples of the ID$_1$ and ID$_2$ derivative generatos. %
    (m-o) shows OOD derivative samples. %
    RMSE: root mean squared error
}
    \label{fig:qualitative}
\end{figure}

Representative samples of the prediction errors for the baseline and DeepLight model can be found in Figure~\ref{fig:qualitative}.
It can be seen that Sobolev training reduced the overall error of the absorbed energy, however we are not able to identify particular structures or features that were more accurately learned by the DeepLight model.
For the derivatives, we observed that the baseline model exhibits the largest error near regions with zero derivative. %
The DeepLight model substantially reduces the error in those regions. %
\par

\section{Discussion}
This study investigated the effect of stochastic Sobolev-training on the performance of an image-to-image surrogate model for light transport. %
To the best of our knowledge, this represents the first application of Sobolev training to a problem of such high dimensionality. %
Our findings show that the DeepLight network achieves improved prediction accuracy when a single directional derivative is incorporated into the loss function for each training sample. %
This makes Sobolev-training of particular interest for surrogate models in situations where the generation of training data is computationally highly expensive, and derivatives can be obtained at little additional cost.\par

We analyzed both ID and OOD samples and observed that Sobolev training reduced errors in the predicted absorbed energy and its derivatives in both datasets. %
Previous studies in lower-dimensional problems have demonstrated that Sobolev training enhanced network generalization and the ability to accurately recover derivatives~\cite{Czarnecki2017,Tsay2021}. %
Now, our present results show that these benefits extend to high-dimensional image-to-image networks as investigated here.\par 

An important future step is to apply the networks presented here to an inverse problem and analyze the effect of Sobolev training on the estimation of optical coefficients and chromophore concentrations. %
A prior study has already demonstrated the advantages of Sobolev training for inverse problems in lower-dimensional models~\cite{Tsay2021}. %
We expect to find similar improvements for reconstructions using the model presented here, motivated by two main arguments. %
Firstly, as network derivatives more closely match the underlying operator derivatives, gradient-based minimizers are likely to produce better optimization steps, leading to faster and more reliable convergence. %
Secondly, with the improved OOD performance observed with the DeepLight model, reconstructions of tissues that are not represented in the training dataset are expected to converge closer to the true absorption coefficients of the tissue. %

Despite the improvements in OOD samples, the DeepLight model still performed best on ID samples. %
To achieve optimal results, simulated training data should therefore match real tissues as closely as possible. %
Although the data generators we used already try to replicate tissue properties, their design remains rudimentary. %
Foundation models based on human anatomy could offer a way to construct more realistic data points, without compromising the necessary generality. %
Additionally, modifying the generators of the derivate directions offers another avenue for future advancements. %
By analyzing the derivative directions encountered in a reconstruction, we could gain valuable insights into constructing a more effective derivative generators tailored for solving the inverse problem.\par

The better performance of the Sobolev-trained model in deeper tissue regions indicates that derivatives are particularly informative for these deeper layers. %
Moving forward, a better understanding of how the information contained in those derivatives affects prediction accuracy could enable more targeted surrogate models for specific applications. %

As the network can reproduce accurate derivatives, it can also be used in tasks involving sensitivity analysis. %
Unlike MoCA, which only supports the computation of forward derivatives, the surrogate model can also efficiently compute adjoint derivatives. %
This capability allows for further applications, such as determining the smallest change in chromophore concentrations that can still be observed given a noise level in the data. %
Answering this kind of question can allow us to investigate if certain diseases or their effects on tissue composition can in principle be captured with MSOT without the need for clinical trials, or what improvements in e.g. detection accuracy would be necessary to do so.\par

In this work, we focused on predicting a 2D projection of absorbed energy due to computational resource constraints. %
Nevertheless, we anticipate that extending the network to volumetric simulations will be a straightforward process. %
An additional limitation lies in the fact that anisotropy is not provided as an explicit input to the network, but is instead represented implicitly in the reduced scattering coefficient. %
While this contains all necessary information for solving light transport in the regime of the diffusion equation, it is in general necessary to consider scattering and anisotropy as independent parameters. %
Although the anisotropy could be trivially added as an additional input channel, reliably learning its effect on absorbed energy independent from scattering could require a substantially larger dataset.\par

In summary, we present a training method for high-dimensional surrogate models intended for solving inverse problems. %
The improved prediction and derivative accuracy for both ID and OOD samples could enable faster and more accurate reconstructions of chromophore concentrations in OptA imaging. %
These developments would greatly improve the clinical utility of this modality and support further downstream tasks that depend on accurate reconstructions. %
In general, Sobolev-trained surrogate models have the potential to better utilize the immense speed-up available with neural networks in conventional variational inversion methods in a variety of problems. %

\appendices
\section*{Acknowledgment}
The authors would like to thank Lukas Immanuel Scheel-Platz, Suhanyaa Nitkunanantharayah, Christoph Dehner, Sarah Franceschin, Sarkis Ter Martirosyan, Maximilian Bader and Guillaume Zahnd for the fruitful discussions and valuable inputs, as well as Serene Lee for her assistance with the preparation of the manuscript.
\section*{Conflict of Interest}
V.N. is a founder and equity owner of Maurus OY, sThesis GmbH, Spear UG, Biosense Innovations P.C. and I3 Inc.
\bibliographystyle{ieeetr}
\bibliography{./MachineLearning,./Optoacoustics,./ProbabilityTheory}

\end{document}